\documentclass[10pt,twocolumn,letterpaper]{article}

\usepackage[pagenumbers]{cvpr} 

\usepackage{graphicx}
\usepackage{amsmath}
\usepackage{amssymb}
\usepackage{booktabs}

\usepackage[pagebackref,breaklinks,colorlinks]{hyperref}

\usepackage[capitalize]{cleveref}
\usepackage[utf8]{inputenc}
\crefname{section}{Sec.}{Secs.}
\Crefname{section}{Section}{Sections}
\Crefname{table}{Table}{Tables}
\crefname{table}{Tab.}{Tabs.}

\def\cvprPaperID{43} 
\def\confName{CVPR}
\def\confYear{2022}

\begin{document}

\title{Towards efficient feature sharing in MIMO architectures}

\author{Rémy Sun$^{1,3,}$\thanks{Corresponding author \hfill mailto:
    remy.sun@isir.upmc.fr} \hspace{.5cm} Alexandre Ramé$^1$ \hspace{.5cm}
  Clément Masson$^3$ \hspace{.5cm} Nicolas Thome$^2$
  \hspace{.5cm} Matthieu Cord$^{1,4}$ \and
  \\
  $^1$MLIA, ISIR, Sorbonne Université \and
  $^2$Vertigo, CEDRIC, Conservatoire National des Arts et Métiers \and
  $^3$Thales Land and Air Systems, Elancourt, France \and
  $^4$Valeo.ai
}

\maketitle

\begin{abstract}
Multi-input multi-output architectures propose to train multiple subnetworks
within one base network and then average the subnetwork predictions to benefit
from ensembling for free. Despite some relative success, these architectures are
wasteful in their use of parameters. Indeed, we highlight in this paper that the
learned subnetwork fail to share even generic features which limits their
applicability on smaller mobile and AR/VR devices. We posit this behavior stems
from an ill-posed part of the multi-input multi-output framework. To solve this
issue, we propose a novel unmixing step in MIMO architectures that allows
subnetworks to properly share features. Preliminary experiments on CIFAR 100
show our adjustments allow feature sharing and improve model performance for
small architectures.
\end{abstract}

\section{Introduction}
\label{sec:intro}


The last decade has seen large deep architectures take over many machine
learning domains \cite{NIPS2012_c399862d, preacthe}
previously solved by more traditional algorithms. As such, deep learning
has become ever more present in practical applications. It is therefore now especially important to find ways to maximize model performance \cite{zhang2018mixup, Shanmugam_2021_ICCV}.


A well known way to obtain better performances given already trained models is
to ensemble the predictions given by multiple models
\cite{lakshminarayanan2016simple}. Indeed, predictions from independently
trained models have been shown to complement each other such that the aggregated
predictions largely outperform the individual model predictions on a test set.


Unfortunately, this increase in performance comes at the cost of dramatically
increased overhead : to ensemble multiple models, one must have access to
multiple trained models \cite{lakshminarayanan2016simple, hansen1990neural}.
This is an untenable cost in many real world applications where networks must
fit on tiny embedded chips in mobile and AR/VR devices. Significant emphasis has therefore been put in the
ensembling literature on finding ways to minimize the inherent cost of
ensembling, typically through some degree of parameter sharing between models
\cite{lee2015m, wen2019batchensemble}.

\begin{figure}
  \centering
  \begin{subfigure}{\linewidth}
    \centering
    \includegraphics[width=\linewidth]{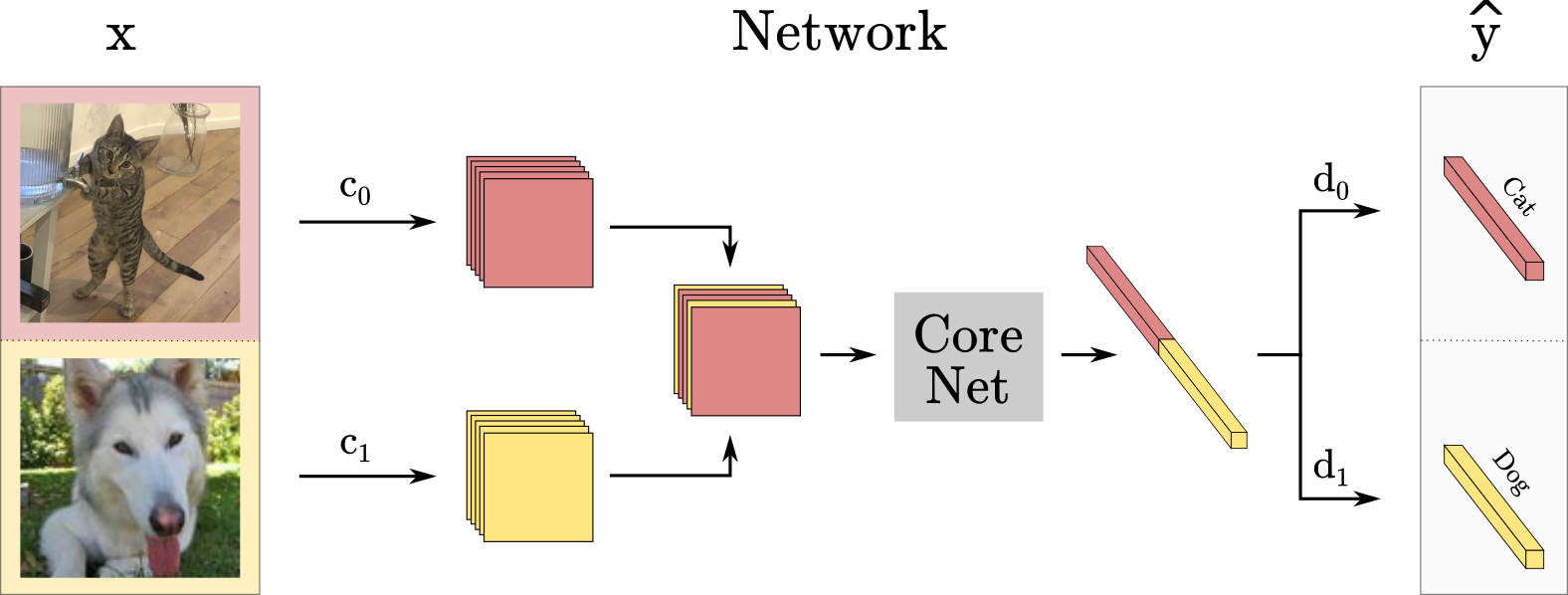}
    \caption{Previous MIMO models train subnetworks that fail to share
      features.}
    \label{fig:intro_mimo}
  \end{subfigure}
  \begin{subfigure}{\linewidth}
    \vspace{.25cm}
    \centering
    \includegraphics[width=\linewidth]{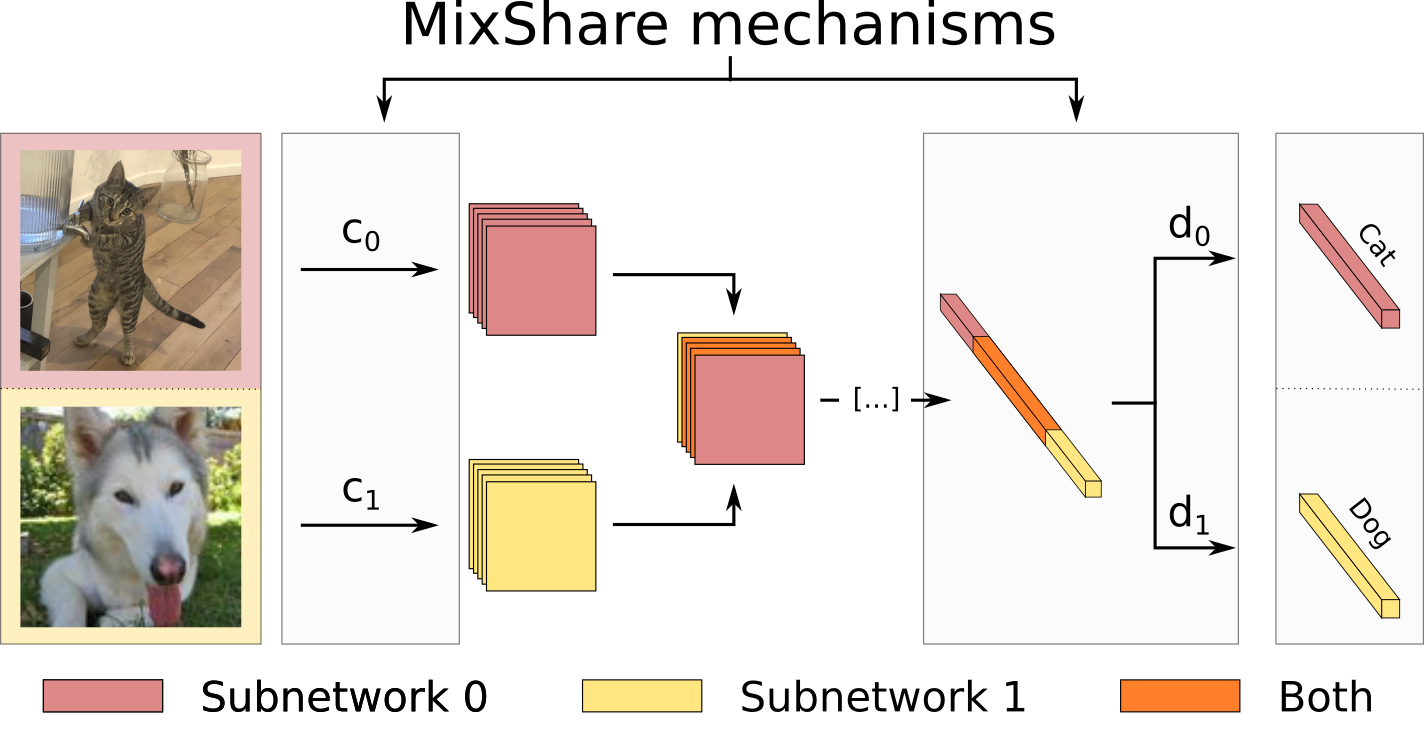}
    \caption{MixShare introduces mechanisms that allow feature sharing.}
    \label{fig:intro_sep}
  \end{subfigure}
  \caption{The lack of feature sharing in MIMO models is a missed opportunity
    for parameter efficiency on smaller networks. Mixshare introduces unmixing to allow feature sharing among
subnetworks in multi-input multi-output models, improving them on smaller
networks previous methods struggle on.}
  \label{fig:intro}
\end{figure}


Multi-input multi-output (MIMO) strategies \cite{havasi2021training,
rame2021mixmo} provide an interesting solution to this conundrum by ensembling
for virtually free. Through their multiple inputs and outputs, MIMO
frameworks train independent subnetworks within a base network. Thanks to the
sparse nature of large neural networks \cite{malach2020proving}, the resulting
subnetworks yield strong and diverse predictions that can be ensembled.
As shown on Fig.~\ref{fig:intro_mimo} with $M=2$, the $M$
inputs are embedded by $M$
subnetworks with no structural differences. Thus, we have $M$ (inputs, labels)
pairs in training: $\{(x_i, y_i)\}_{0 \leq i < M}$. More precisely, \textbf{$M$ images are fed to the network
at once}. The $M$ inputs are encoded by $M$ distinct convolutional layers
$\{c_i\}_{0 \leq i < M}$ into a shared latent space before being aggregated
(either through summation \cite{havasi2021training} or more complex mixing
operations \cite{rame2021mixmo}). This representation is then processed by the
core network into a single feature vector, which is classified by $M$ dense
layers $\{d_i\}_{0 \leq i < M}$. Diverse subnetworks appear
as $d_i$ learns to classify $y_i$ from input $x_i$. At inference, we can
ensemble $M$ predictions by feeding the same image $M$ times to the model.

MixMo \cite{rame2021mixmo} has however recently highlighted significant limitations of such
architectures: multi-input multi-output architectures require large base models
and struggle to fit more than 2 subnetworks. Indeed, \cite{rame2021mixmo} shows a significant drop in performance
on CIFAR 100 when going from 2 subnetworks to 4 subnetworks.


This scaling issue is explained by analyzing the features inside the network, as
we show at the beginning of this paper by extending \cite{havasi2021training,
rame2021mixmo}'s study of subnetwork behavior. Our analysis shows that the
aforementioned scaling issues stem from how subnetworks share no features
in the base network (see Fig.~\ref{fig:intro_mimo}): each channel or feature is
almost exclusively used by one subnetwork. As such, we can explain the
scaling issue since each additional subnetwork significantly reduces the
effective size of the individual subnetworks. Beyond causing issues on smaller
architectures or harder datasets, this leads to very wasteful use of network
parameters. This is especially unfortunate as the subnetworks could at the very
least share generic features in the first layers. We see this as a missed
opportunity, one that can significantly improve multi-input multi-output models'
applicability to real world settings like mobile devices.


In this paper, we first carefully study in Sec.~\ref{sec:why} how subnetworks
use the base network's features. After showing the lack of feature sharing, we
discuss the impact of this on parameter efficiency and model performance.
Secondly, we propose Mixshare in Sec.~\ref{sec:mixshare} to address the issues
preventing feature sharing (see Fig.~\ref{fig:intro_sep}). In particular, we
introduce a novel unmixing mechanism (Sec.~\ref{sec:unmix}) to allow sharing and
discuss in Sec.~\ref{sec:init} how proper network initialization is necessary to
improve model performance.

\section{MIMO Subnetworks do not share features}
\label{sec:why}

In this section, we strive to pinpoint the cause of multi-input multi-output
architectures' scaling issues. To this end, we consider the following question:
how do subnetworks behave in multi-input multi-output architectures ?

\begin{figure}[t]
  \centering
  \includegraphics[width=.95\linewidth]{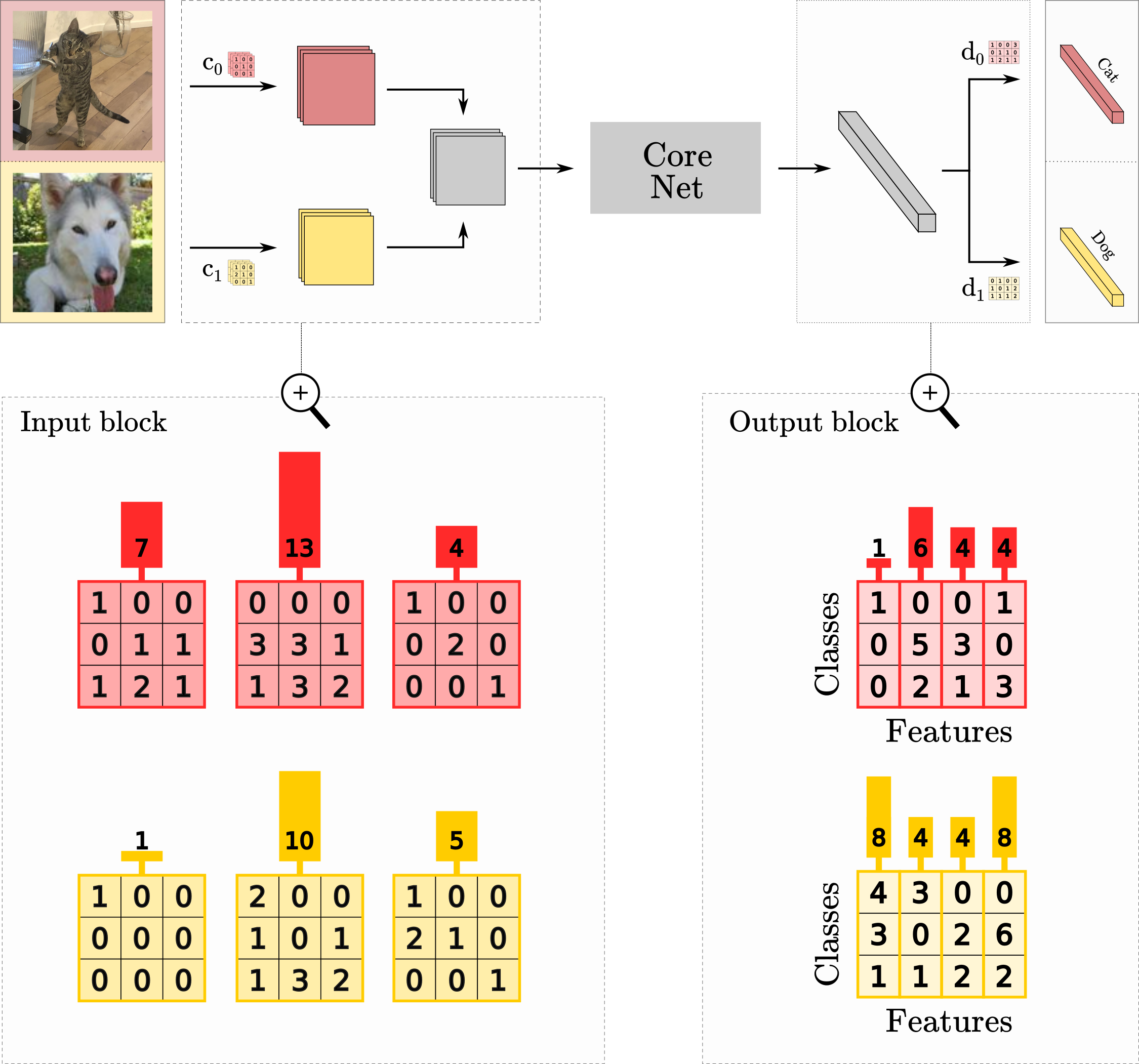}

  \caption{We study the influence of features in the input and output block on
    the subnetworks ($L_1$ norm figured by bars). For the input block, we
    consider the $L_1$ norm of the feature kernels for the relevant encoders. For the output block,
  we look at the $L_1$ norm of columns in the classifier weights matrices. On the
  figure, this shows us the first input block kernel is mostly used by the first
  (red) subnetwork (7 vs. 1). Similarly, the first feature of the output block proves
 important only to the first (red) classifier.}
  \label{fig:study}
\end{figure}

Following \cite{rame2021mixmo}, we check how the inputs are organized in the $C$ feature
maps of the mixing space by considering the $L_1$ norm of the $C$ encoder kernels for
each subnetwork (see Fig.~\ref{fig:study}). This tells us whether a feature map
contains more information about one input, and we can visualize which maps
are used by which subnetwork through histograms $h_0$ and $h_1$ of feature
influence for each subnetwork. Quantitatively, we can approximate the feature
sharing rate through the ratio of $\frac{\min(h_0,h_1)}{\max(h_0,h_1)}$.
In the same spirit, we consider the $L_1$ norm of the columns of classifier weight
matrices to quantify the importance of each feature to each classifier.

We conduct our study on a WideResNet-28-2 \cite{BMVC2016_87} using the more realistic batch
repetition 2 setting from \cite{rame2021mixmo} on the CIFAR 100 dataset
\cite{krizhevsky2009learning} (see Appendix). We
choose to consider this situation as it perfectly showcases the
issues encountered by MIMO methods on smaller architectures. 
To complement this, we also show results on the slightly larger WideResNet-28-5
later on in the paper.

\begin{figure}
  \centering
  \begin{subfigure}{0.45\linewidth}
    \includegraphics[width=\linewidth]{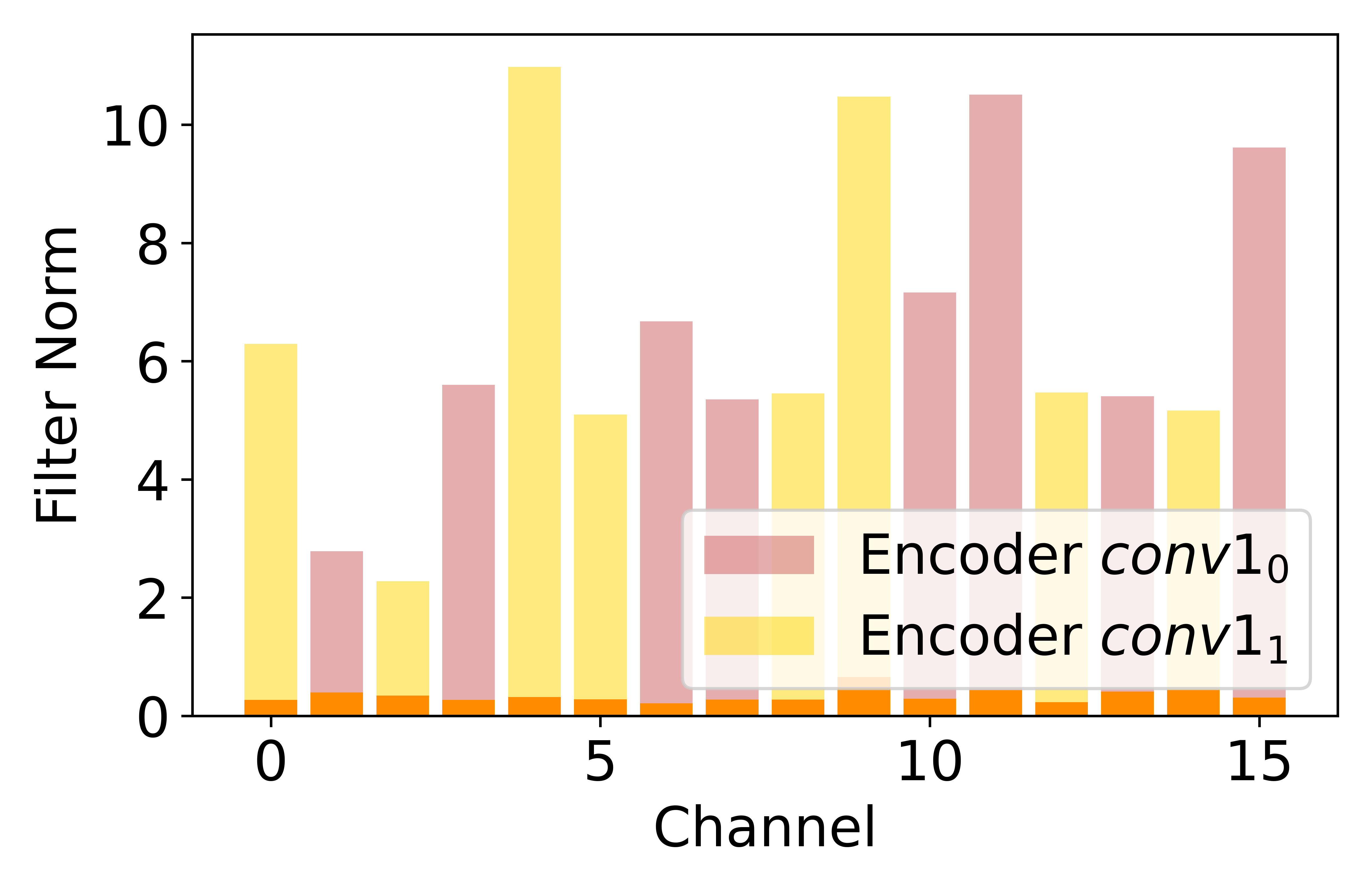}
    \caption{Encoder kernels $L_1$ norm.}
    \label{fig:short-a}
  \end{subfigure}
  \hfill
  \begin{subfigure}{0.45\linewidth}
    \includegraphics[width=\linewidth]{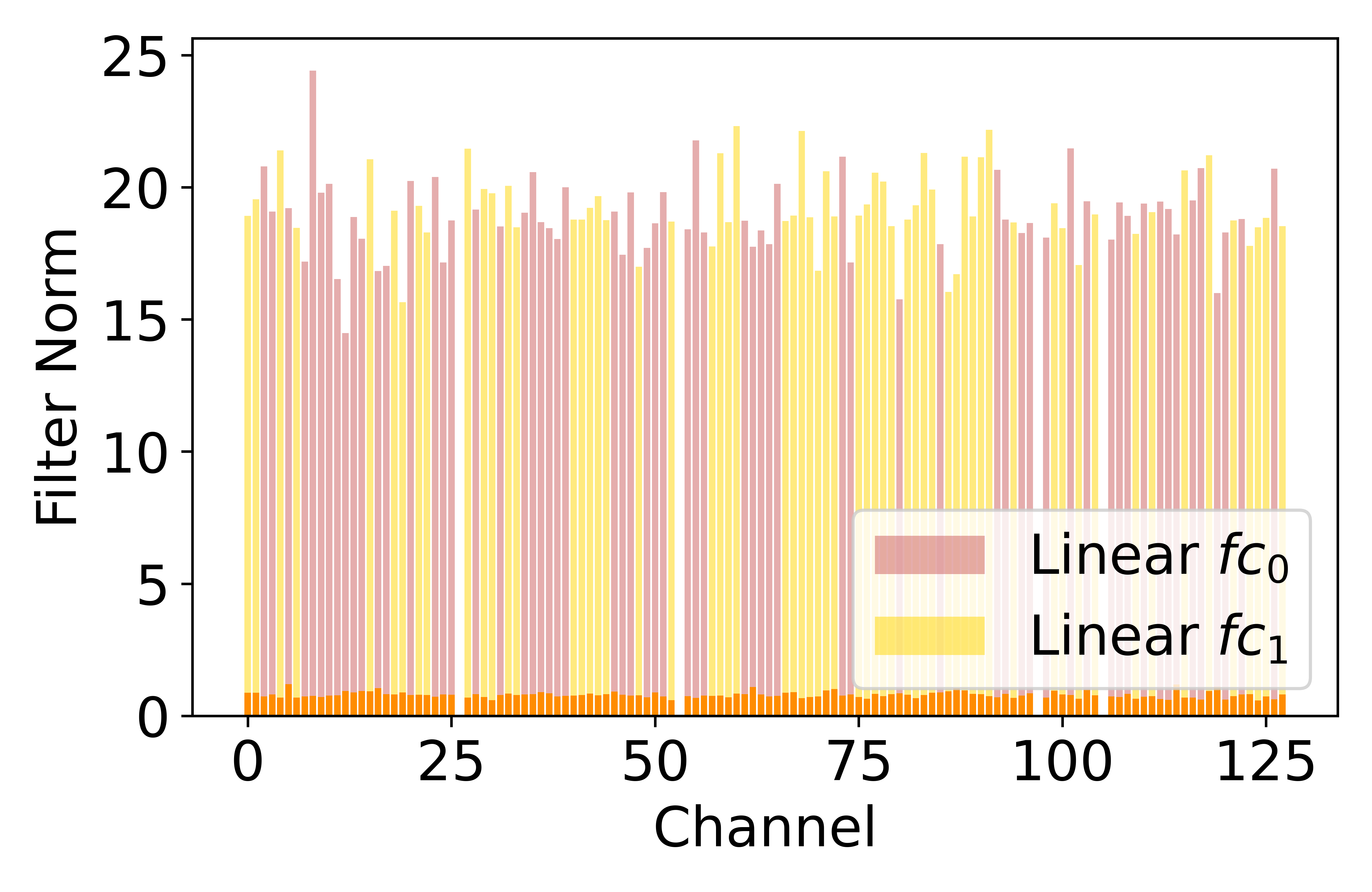}
    \caption{Classifier columns $L_1$ norm.}
    \label{fig:short-b}
  \end{subfigure}
  \caption{Features are used by one subnetwork or the other, never both at the
    same time: the overlap (orange) is very low.}
  \label{fig:sep}
\end{figure}

Fig.~\ref{fig:sep} shows the subnetworks are fully independent in the core
network: each channel in the input block encodes information about only
one input, as the corresponding kernel of the other encoder's $L_1$ is very low. A similar behavior is observed in the output block, and further
analysis of input influence on intermediary feature maps shows this behavior
remains consistent within the network (See Appendix).


Multi-input multi-output architectures' scaling issues become much easier to
understand in light of this: the amount of weights available to each of the
underlying subnetworks decreases quadratically with the number of subnetwork.
Indeed, since feature maps of different subnetworks cannot communicate, only
$\frac{1}{M}$ weights can be non-zero. This fraction of non-zero weights must
then be distributed between the $M$ subnetworks. Furthermore, the subnetworks
likely extract similar generic features, at least in the first layers. Since the
subnetworks share no features, this means those features are unnecessarily
replicated for each subnetwork.


This is not wholly surprising or undesirable behavior as MIMO strives to
train $M$ independent subnetworks to obtain diverse ensembles. By
avoiding overlap between subnetworks, the subnetworks act as a standard
ensemble of smaller models, with the base model size acting as hard cap on the
number of the parameters used by the ensemble.


While it is true not sharing any features ensures
subnetworks' independence, it seems unnecessary. Indeed, the
subnetworks are highly unlikely to extract completely different features. As such,
subnetworks should benefit from sharing features at least in the early layers
even if the classifier still consider fairly different features.


At first blush, nothing in the MIMO
training protocol explicitly requires the subnetworks not share any
features. Why do the subnetworks avoid sharing features ? How could we
encourage them to share some parameters ?

\begin{figure}
  \centering
  \includegraphics[width=\linewidth]{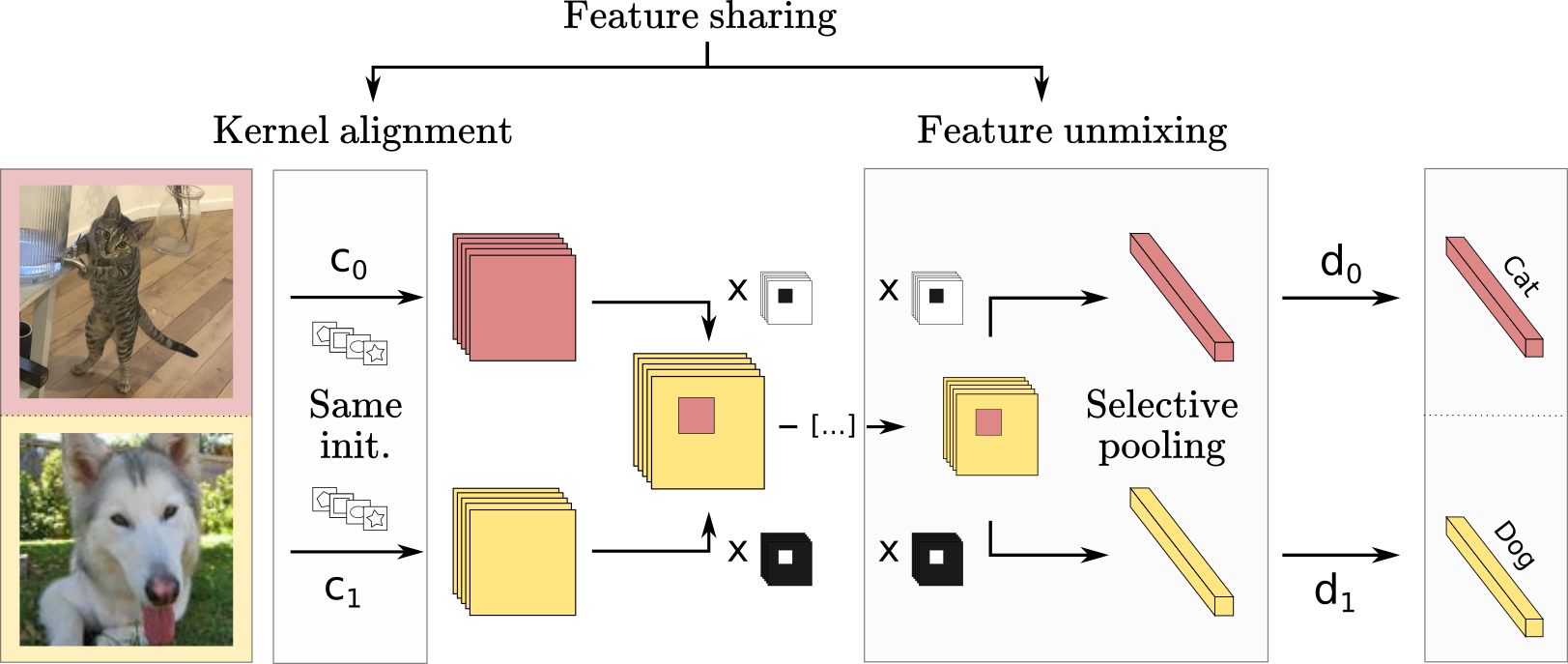}
  \caption{Two steps are necessary to allow feature sharing in networks:
     1) Ensure the subnetworks share a ``common language'' by initializing the
     convolutional encoders to be close to each other. 2) Extract descriptions of each input from model features.}
  \label{fig:unmixing}
\end{figure}

\section{How can subnetworks share features ?}
\label{sec:mixshare}

We discuss here the obstacles preventing feature sharing in multi-input
multi-output architectures, and propose solutions to correct this behavior.

\subsection{Unmixing: extracting features for each input}
\label{sec:unmix}


We build upon an intuition put forth in MixMo \cite{rame2021mixmo}: the lack of
feature sharing is caused by the need for individual classifier at the end of
the network to extract class information for one input specifically. Indeed, the
$M$ classifiers have access to the exact same set of extracted features. If two
classifiers use the same feature, that feature needs to describe the state of
two different inputs. This is an issue when one accounts for the fact inputs are
in fact drawn independently and there can therefore be no meaningful feature
describing the state of two inputs simultaneously.


\begin{figure}
  \centering
  \begin{subfigure}{0.45\linewidth}
    \includegraphics[width=\linewidth]{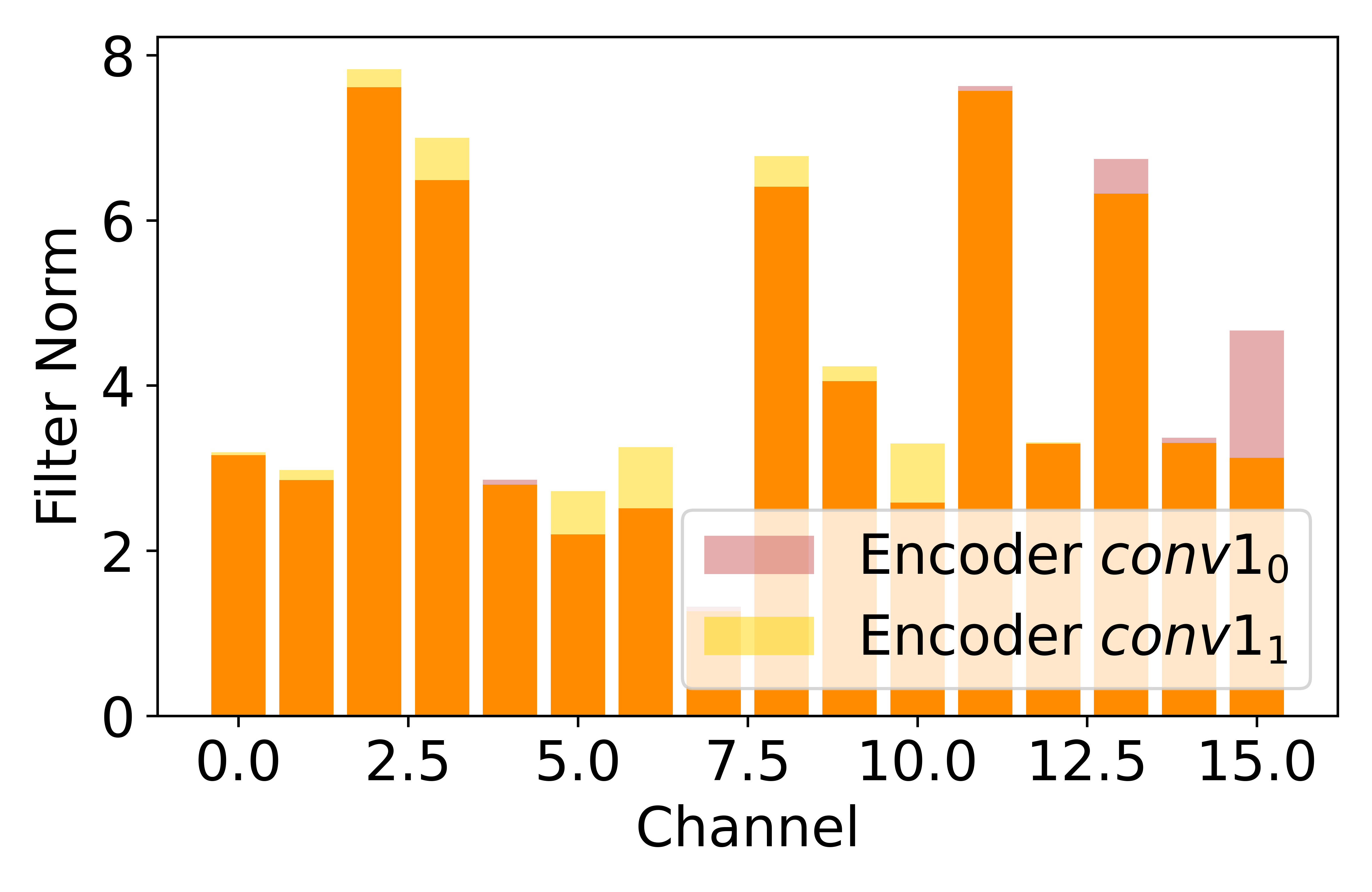}
    \caption{Encoder kernels $L_1$ norm.}
    \label{fig:short-a}
  \end{subfigure}
  \hfill
  \begin{subfigure}{0.45\linewidth}
    \includegraphics[width=\linewidth]{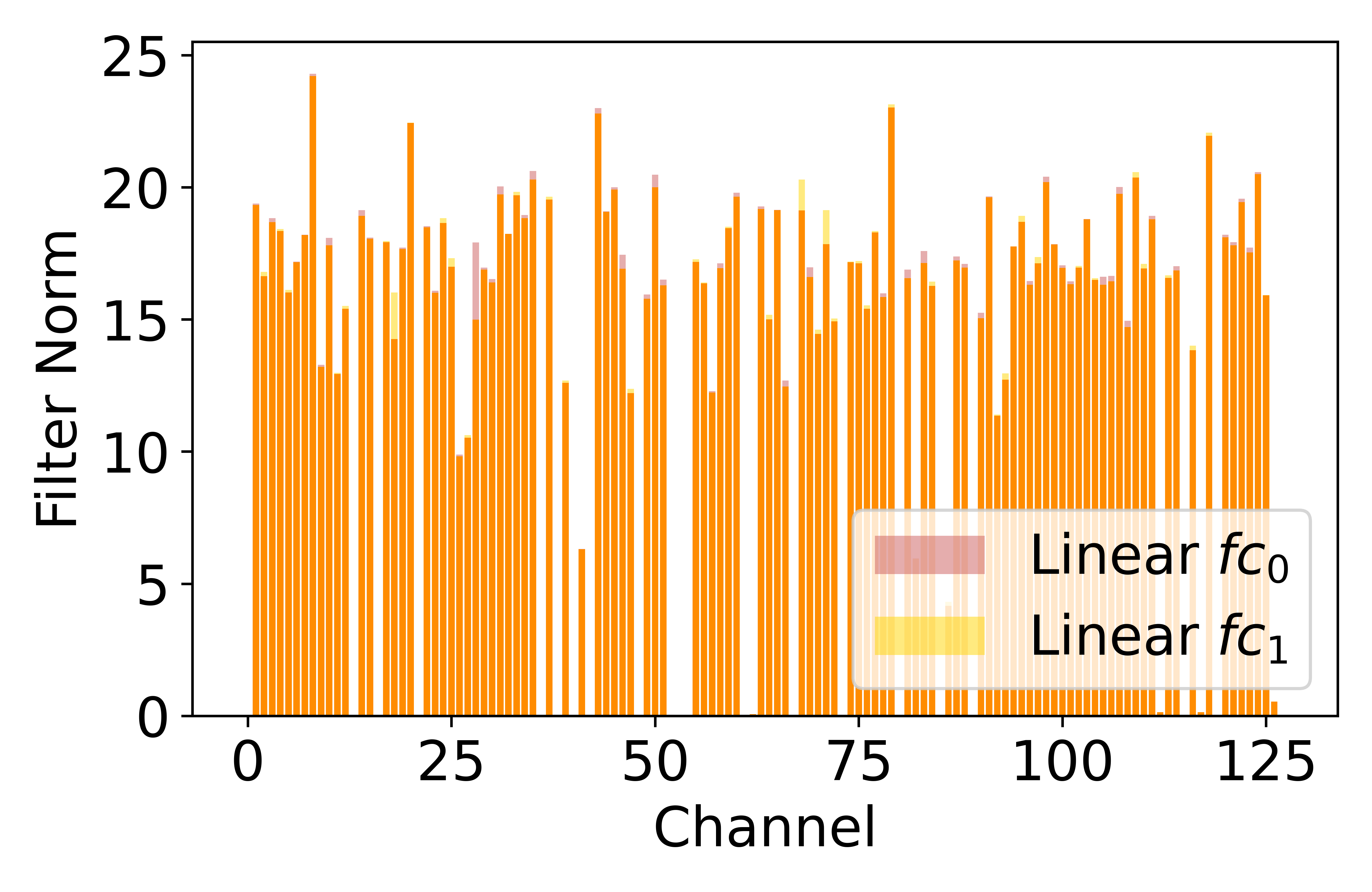}
    \caption{Classifier columns $L_1$ norm.}
    \label{fig:short-b}
  \end{subfigure}
  \caption{Applying unmixing leads to features being used by both subnetworks:
    the overlap (orange) is very high.}
  \label{fig:featshare}
\end{figure}


\begin{table*}
  \centering
  \scalebox{0.875}{
  \begin{tabular}{@{}lcccccc@{}}
    \toprule
    Method & \multicolumn{3}{c}{28-2} &  \multicolumn{3}{c}{28-5} \\ \cmidrule{2-4} \cmidrule{5-7}
           & Acc. ens. & Acc. Ind. &  Classifier share rate &  Acc. ens. & Acc. Ind. & Classifier share rate \\
    \midrule
    MixMo & $74.8 \pm 0.2$ & $71.6 \pm 0.2$ & $4.9 \pm 0.4$ & $81.9 \pm 0.1$ & $79.4 \pm 0.2$ & $8.2 \pm 0.1$ \\
    MixMo + Unmix & $69.7 \pm 15.4$ & $69.7 \pm 15.4$ & $99.1 \pm 0.4$ & $79.4 \pm 2.7$ & $79.4 \pm 2.7$ & $98.8 \pm 0.7$  \\
    MixMo + Unmix + kernel init. & $\mathbf{79.0 \pm 0.1}$ &  $79.0 \pm 0.1$ & $99.4 \pm 0.1$ & $82.1 \pm 0.2$ & $82.1 \pm 0.2$ & $99.3 \pm 0.1$ \\
    \midrule
    MixShare (\textbf{partial} on 25\% features) & $73.3 \pm 0.5$ & $71.5 \pm 1.0$ & $60.6 \pm 6.5$ & $79.9 \pm 0.3$ & $78.8 \pm 0.4$ & $60.0 \pm 2.3$  \\
    MixShare (\textbf{fadeout} to 100 epochs) & $\mathbf{79.0 \pm 0.1}$ & $76.7 \pm 1.3$ & $64.4 \pm 6.0$ & $\mathbf{82.4 \pm 0.3}$ & $81.6 \pm 0.5$ & $62.7 \pm 8.3$ \\
    \bottomrule
  \end{tabular}
  }
  \caption{Overall ensemble accuracy (\%), average subnetwork accuracy (\%) and
    classifier (output block) sharing rate (see Sec.~\ref{sec:why})
    for MixMo variants,
    $mean \pm std$ reported over 3 runs. Mixshare with fadeout unmixing
    yields both strong
    individual models and ensemble gains.}
  \label{tab:results}
\end{table*}


Let us now consider how the classifiers should ideally behave on shared
features. Since each classifier is paired to one of the input pathways, they
should be able to extract two different interpretations of the shared features
that still encode the same functional information (see Fig.~\ref{fig:unmixing}). For instance, the
shared feature should encode for the presence of flowers but each
classifier should be able to infer from the feature whether its personal input
contains flowers.


While this is not the case in traditional CNNs, MixMo \cite{rame2021mixmo}
introduces a modification to the seminal MIMO architecture that causes feature
maps to encode information about the different inputs separately. Indeed, since
MixMo mixes inputs according to some binary mixing augmentation scheme (typically CutMix \cite{yun2019cutmix}),
each pixel on the final feature maps encodes information about one of the inputs.


This is fortunate as it provides us with a fairly natural solution: unmixing.
\textbf{Unmixing} (illustrated in Fig.~\ref{fig:unmixing}) recycles the binary
masks generated for input mixing in order to filter the feature maps so that
only information relevant to a specific input is contained in the unmixed
version. This way, a single feature map can describe each of the inputs.

Fig.~\ref{fig:featshare} shows that applying unmixing causes the subnetworks to
share features, both in the input and output block. In fact, every
feature in the unmixed model is used by all subnetworks which proves unmixing
indeed solves the core obstacle to feature sharing in MIMO networks.


Introducing unmixing however leads to unstable and generally worse performance
as seen in Tab.~\ref{tab:results}. Crucially, even individual subnetwork
accuracy suffers from unmixing which suggests an underlying issue.

\subsection{Aligning encoder kernels to allow efficient feature sharing}
\label{sec:init}


Intuitively, feature sharing should at the very least lead to higher individual
subnetwork accuracy as the subnetworks use more parameters. As such, we now
investigate why unmixing degrades performance so dramatically.


By extracting multiple possible interpretations of a single feature, unmixing
introduces a new problem in the model. Indeed, we need our interpretations of
the same feature to encode the same functional characteristics (e.g. flower
detection). The issue is that a randomly initialized multi-input multi-output
network typically leads to having multiple interpretations of the same feature.


Indeed, the encoders computing the mixed representations are very
different. For an input feature, the mixed feature map could
contain information about horizontal borders on input 1 and vertical borders on
input 2. As such, there is no consistent interpretation for our mixed features.


We can unify the interpretation of unmixed features at the start by simply
\textbf{aligning the kernels of the encoders}. Indeed, as long as each feature encodes
the same sort of information for each encoder, there should be no ambiguity
introduced by the unmixing process. 


Tab.~\ref{tab:results} shows that fixing the initialization scheme of
the encoders to the same value does indeed lead the model to outperform normal mixmo models. 

\subsection{Towards partial feature sharing}


While proper unmixing does allow feature sharing in multi-input multi-output
networks, Tab.~\ref{tab:results} and Fig.~\ref{fig:featshare} show it leads
to subnetworks sharing all features: the subnetworks are identical. This is even less desirable
than fully separated subnetworks as it makes ensembling pointless
\cite{pang2019improving, rame2021dice}.


Ideally, subnetworks would share some parameters but still remain distinct functionally.
This way, we would be able to strike a compromise between fully separated
and fully shared subnetworks. The issue with this however, is
that removing obstacles to feature sharing makes it unnecessary for subnetworks
to separate in any way.


In this preliminary work, we discuss two solutions: \textbf{partial
unmixing} and \textbf{fadeout unmixing}. \textbf{Partial
unmixing} is a straightforward solution where we only apply unmixing to a fixed subset
of the final feature maps (e.g. 25\%). In \textbf{Fadeout unmixing} we start training the
network with proper unmixing but progressively reduce the strength of unmixing
so that there is no unmixing towards the end of the procedure. For instance, we
use the unmixing mask $M + r (1-M)$ (instead of $M$) with $r=\min(1, epoch/100)$
if we want to stop unmixing by epoch 100. As such, fadeout unmixing initializes
the network in a shared state and progressively pushes the subnetworks to
develop independent features.


We now propose the full MixShare framework by combining proper kernel
initialization and partial/fadeout unmixing along with slight
adjustments to standard MIMO procedures like input repetition
\cite{havasi2021training} and loss balancing \cite{rame2021mixmo} (see Appendix).
Tab.~\ref{tab:results} shows that both MixShare variants succeed in causing partial
feature sharing. \textbf{Partial} fails to train strong individual subnetworks, but
still showcases ensemble benefits. \textbf{Fadeout} on the other
hand leads to strong performances and retains significant ensembling benefits on
medium sized networks like a WideResNet 28-5.

\section{Conclusion}

We have shown multi-input multi-output models induce fully
separated subnetworks because of a difficulty in matching outputs to inputs for
the neural network. We have proposed an unmixing mechanism and encoder initialization for MixMo
\cite{rame2021mixmo} architectures and demonstrated it
allows multi-input multi-output architectures to share features. Our preliminary experiments show this
corrected architecture outperforms standard multi-input multi-output
architectures on smaller networks with a proper unmixing scheme. We hope that by
highlighting the main issue at the crux of these architectures' inefficiency,
our work will lead to further research on MIMO architectures that will lead to
their deployment smaller mobile and AR/VR devices.


\paragraph{Acknowledgments}

This work was conducted using HPC resources from GENCI–IDRIS (Grant
2021-AD011013208), and under a CIFRE grant between
Thales Land and Air Systems and Sorbonne University.

{\small
\bibliographystyle{ieee_fullname}
\bibliography{egbib}
}

\newpage

\appendix

\section*{Appendix}

\section{Experimental details}

The code for this work was directly adapted from the official MixMo
\cite{rame2021mixmo} codebase: \url{https://github.com/alexrame/mixmo-pytorch}.

We followed similar experimental settings on CIFAR 100 as MixMo
\cite{rame2021mixmo} and present here the adapted setting description:

We used standard architecture WRN-28-$w$, with a focus on $w=2$. We re-use the
\textbf{hyper-parameters} configuration from MIMO \cite{havasi2021training} with
batch repetition 2 (bar2). The optimizer is SGD with learning rate of
$\frac{0.1}{b}\times\frac{ \text{batch-size}}{128}$, batch size $64$, linear
warmup over 1 epoch, decay rate 0.1 at steps $\{75, 150, 225\}$, $l_2$
regularization 3e-4. We follow standard MSDA practices \cite{yun2019cutmix} and
set the maximum number of epochs to $300$. Our experiments ran on a single
NVIDIA 12Go-TITAN X Pascal GPU.

All experiments were run three times on three fixed seeds from the same version
of the codebase. Qualitative results presented in Fig.~\ref{fig:sep} and
Fig.~\ref{fig:featshare} are obtained by visualizing results for the first
set of random seeds. Quantitative results presented in Tab.~\ref{tab:results}
are given in the form of $mean \pm std$ over the three runs.

\section{Complementary adjustments to MIMO procedures in MixShare}

MIMO methods use a number of auxiliary procedures to train strong subnetworks.
However, as MixShare differs significantly from standard MIMO frameworks, it
does not use these frameworks to the same extent.

\paragraph{CutMix probability in the input block} MixMo \cite{rame2021mixmo}
only uses cutmix mixing in its input block about half the time, using a basic
summing operation on the two encoded inputs the rest of the time. This is
because the model will use a summing operation at test time. Therefore, the use
of cutmix at training induces an strong train/test gap that needs to be bridged
by the use of summing during training.

We cannot afford to use summing half the time as unmixing relies on the use of
cutmix in the input block. However, since our two encoders are very similar (due
to our kernel alignment), cutmix and summing (or averaging) behave very
similarly and the train/test gap is therefore minimal.

\paragraph{Input Repetition} A slight train/test gap still remains however since
the model is rarely presented the same image as input to both subnetworks at
training time. We solve this by reprising a procedure introduced in the seminal
MIMO paper \cite{havasi2021training}: input repetition. In our case, we ensure
10\% of inputs of our batches are made of repetition of the same image during
training.

\paragraph{Loss rebalancing} MixMo \cite{rame2021mixmo} introduced a
re-weighting function of the subnetwork training losses that rescales the mixing
ratios used in the inputs block. These ratios are rescaled to be less lopsided
(closer to an even 50/50 split) before being applied to their relevant
subnetwork losses. This rescaling is necessary as it ensures all parameters
receive sufficient training signal.

We however find in our experiments it is more beneficial to do away with this
re-balancing and keep the original mixing ratios, which we explain by the large
amount of features shared between subnetworks. Since features are shared, we do
not need to worry about some features receiving too little training signal.

\section{A more nuanced discussion on kernel alignment}

While MixShare uses the exact same initialization of the encoder kernels for
simplicity, it is interesting to note much weaker versions of kernel alignment
are sufficient to obtain similar results.

Indeed, we found in our experiments that initializing the kernels to be simply
co-linear is more than enough to ensure proper feature sharing. In fact, this
leads to the exact same performance as using the same initialization and the
encoder kernels quickly converge to similar values.

This further validates our intuition that MIMO models need a ``common language''
to benefit from sharing features: all that is required is for encoder kernels to
extract the same ``type'' of features.

\section{Analysis of subnetwork features within the core network}

\begin{figure*}[t]
  \centering
  \begin{subfigure}{0.3\linewidth}
    \includegraphics[width=\linewidth]{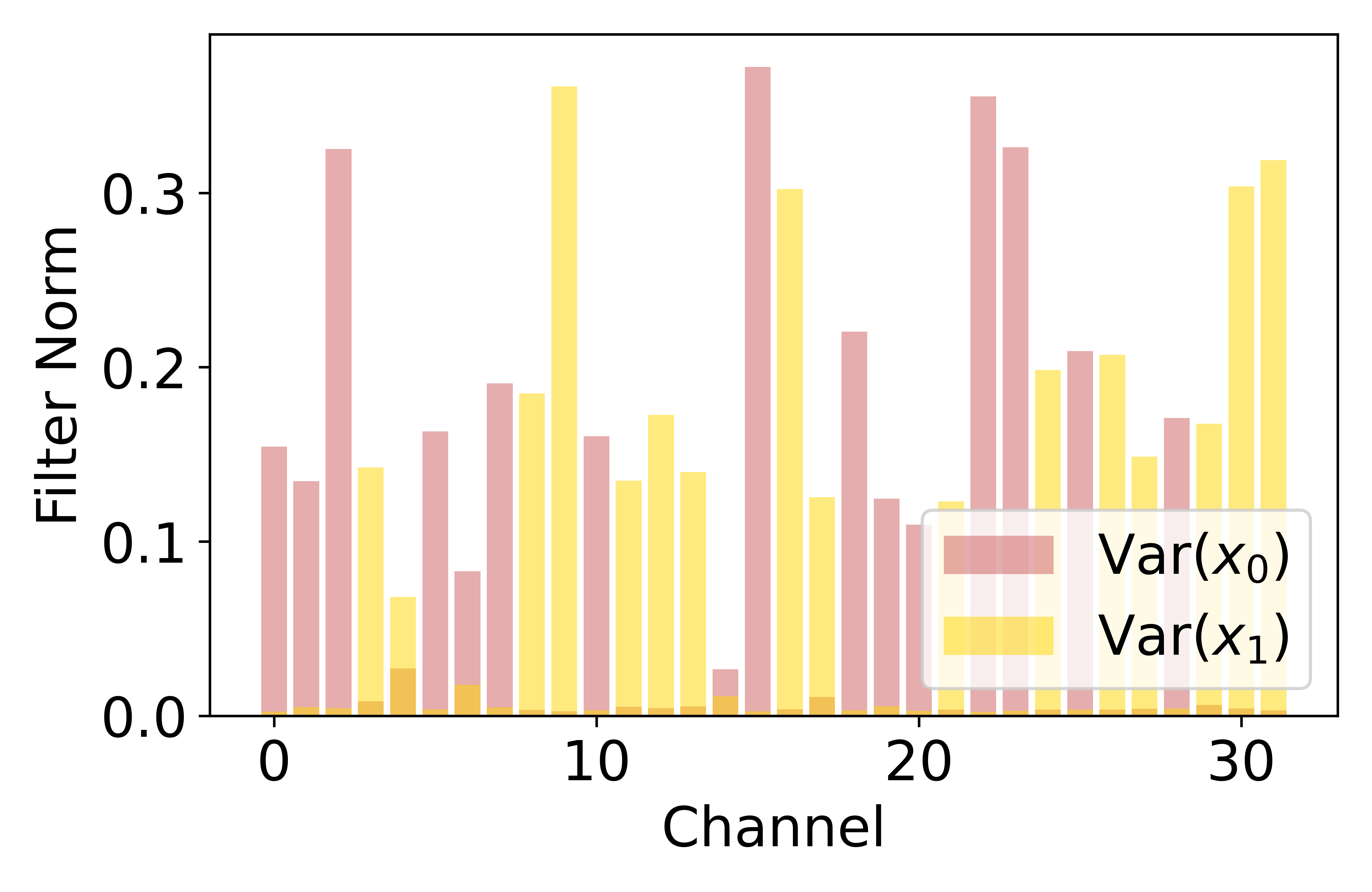}
    \caption{Feature variance after block 1.}
  \end{subfigure}
  \hfill
  \begin{subfigure}{0.3\linewidth}
    \includegraphics[width=\linewidth]{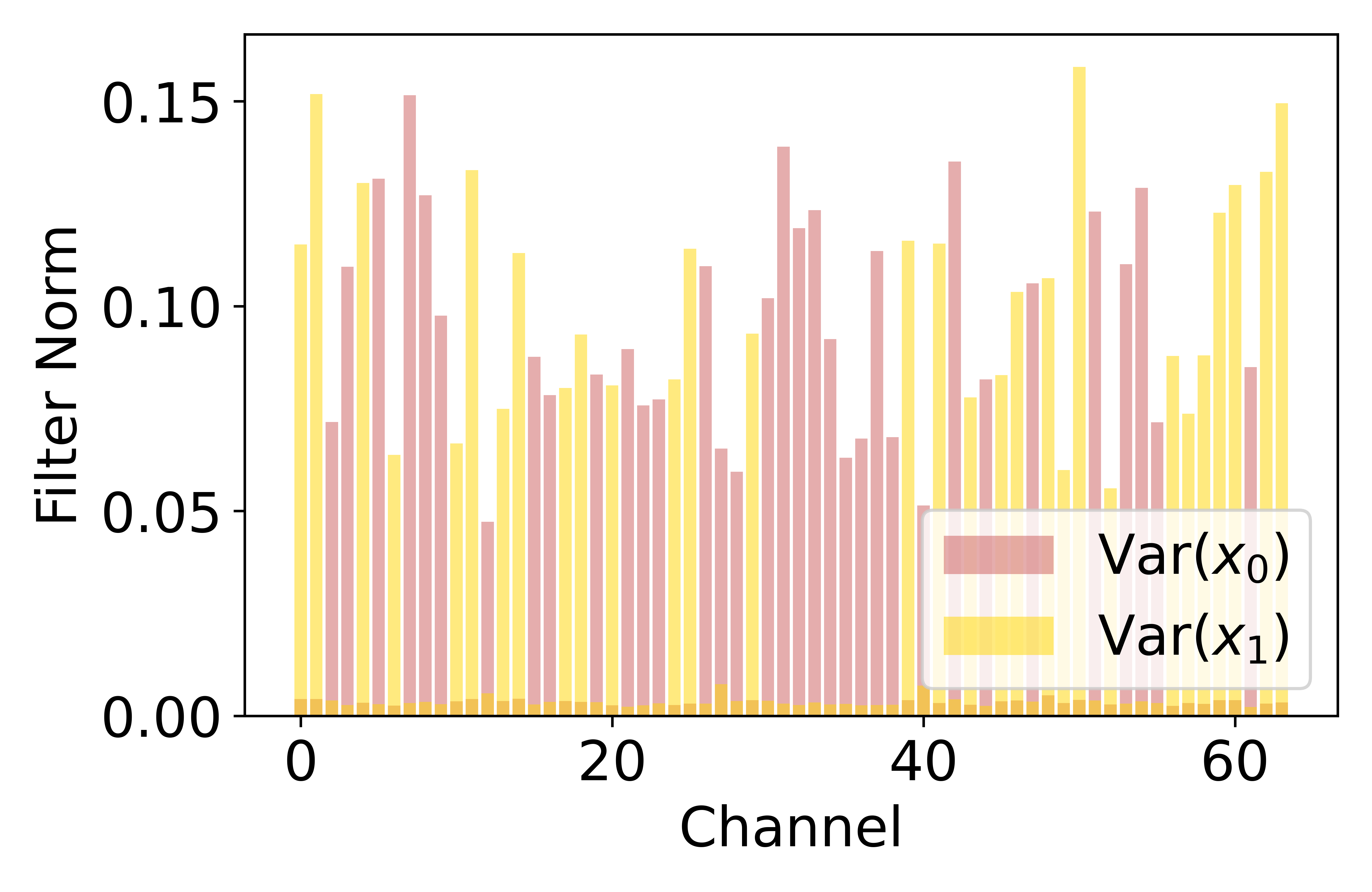}
    \caption{Feature variance after block 2.}
  \end{subfigure}
  \hfill
  \begin{subfigure}{0.3\linewidth}
    \includegraphics[width=\linewidth]{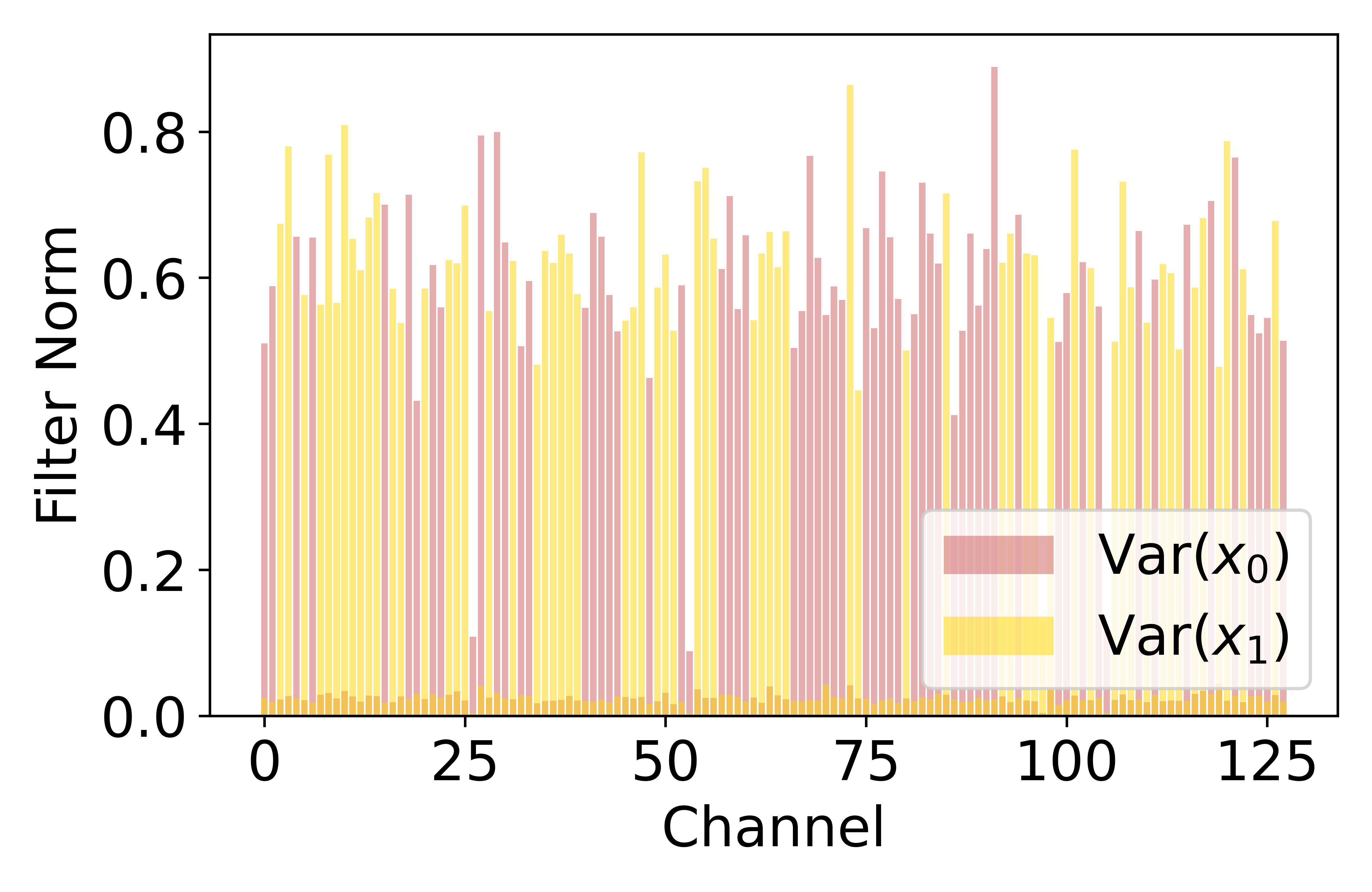}
    \caption{Feature variance after block 3.}
  \end{subfigure}
  \caption{Checking the variance of feature maps w.r.t. the two inputs at
    different levels of the network shows clear separation of features in
    standard multi-input multi-output architectures.}
  \label{fig:var}
\end{figure*}

Sec.~\ref{sec:why} studies what features each subnetwork uses in the input block
and output block of the multi-input multi-output model. Studying the importance
of features within the core networks for each subnetworks is more difficult as
it is not possible to consider the model weights.
Reprising an analysis conducted in the Appendix of \cite{rame2021mixmo}, we
identify the influence of intermediate features on subnetworks with the variance
of the feature with respect to the relevant input.

For the first subnetwork, if we consider the intermediate feature map (at one point in the network $f$) $\mathcal{M}_{int}
= f_{int}(\mathcal{D}_{test},d)$, such that $\mathcal{M}_{int}$ is of shape
$N\times C \times H \times W$ with $\mathcal{D}_{test}$ the test set, $d$ a
fixed input, $N$ the size of the test set, $C$ the number
of intermediate feature maps and $H\times W$ the spatial coordinates. We compute
the importance of each of the $C$ feature map with respect to the first
subnetwork as $Mean(Var(\mathcal{M}_{int}, dim=0), dim=(1,2))$. The importance
of intermediate features for the second subnetwork is obtained similarly by
considering $\mathcal{M}_{int}= f_{int}(d,\mathcal{D}_{test})$.

Fig.~\ref{fig:var} shows the resulting feature importance maps at after each of
the three residual blocks in the core network. As can be observed, the
subnetworks remain consistently separated in the core network.

\end{document}